\documentclass{article}

\PassOptionsToPackage{numbers, compress}{natbib}

\usepackage{graphicx}
\usepackage{booktabs}
\usepackage{makecell}
\usepackage{caption}
\usepackage[dvipsnames]{xcolor}
\usepackage{stmaryrd}

\usepackage[acronym]{glossaries} 

\usepackage{enumitem} 

\usepackage[preprint]{neurips_2020}

\newacronym{lstm}{LSTM}{Long Short-Term Memory}
\newacronym{gru}{GRU}{Gated Recurrent Unit}
\newacronym{convgru}{ConvGRU}{Convolutional GRU}
\newacronym{cnn}{CNN}{convolutional neural network}
\newacronym{convlstm}{ConvLSTM}{Convolutional LSTM}
\newacronym{tsru}{TSRU}{Transformation-based Spatial Recurrent Unit}
\newacronym{ptsru}{PTSRU}{Pixelwise Transformation-based Spatial Recurrent Unit}
\newacronym{stn}{STN}{Spatial Transformer Network}

\usepackage{hyperref}

\newcommand{\enc}{\mathcal{E}}
\newcommand{\gen}{\mathcal{G}}
\newcommand{\disc}{\mathcal{D}}
\newcommand{\tsru}{TSRU}

\newcommand{\mocogan}{MoCoGAN-like}
\newcommand{\trajgru}{TrajGRU}

\newcommand{\dvdganpp}{TrIVD-GAN-FP}

\newcommand{\amended}[1]{{#1}}

\usepackage[utf8]{inputenc} 
\usepackage[T1]{fontenc}    
\usepackage{hyperref}       
\usepackage{url}            
\usepackage{booktabs}       
\usepackage{amsfonts}       
\usepackage{nicefrac}       
\usepackage{microtype}      

\title{Transformation-based Adversarial Video Prediction on Large-Scale Data}

\author{%
  \textbf{Pauline Luc, Aidan Clark, Sander Dieleman, Diego de Las Casas,} \\ 
  \textbf{Yotam Doron, Albin Cassirer, Karen Simonyan} \\
  DeepMind, London, UK \\
  \texttt{\{paulineluc,aidanclark\}@google.com} \\
}

\begin{document}

\maketitle

\begin{abstract}
Recent breakthroughs in adversarial generative modeling have led to models capable of producing video samples of high quality, even on large and complex datasets of real-world video. 
In this work, we focus on the task of video prediction, where given a sequence of frames extracted from a video, the goal is to generate a plausible future sequence. We first improve the state of the art by performing a systematic empirical study of discriminator decompositions and proposing an architecture that yields faster convergence and higher performance than previous approaches. 
We then analyze recurrent units in the generator, and propose a novel recurrent unit which transforms its past hidden state according to predicted motion-like features, and refines it to handle dis-occlusions, scene changes and other complex behavior.
We show that this recurrent unit consistently outperforms previous designs. Our final model leads to a leap in the state-of-the-art performance, obtaining a test set Fr\'echet Video Distance of 25.7, down from 69.2, on the large-scale Kinetics-600 dataset.
\end{abstract}

\section{Introduction}

The ability to anticipate future events is a crucial component of intelligence. 
Video prediction, the task of generating a plausible future sequence of frames given initial conditioning frames, has been increasingly studied as a proxy task towards developing this ability, as video is a modality that is cheap to acquire, available in abundant quantities and extremely rich in information.
It has been shown to be useful for representation learning~\citep{srivastava15unsupervised,walker2016eccv,vondrick2016generating} and in the context of reinforcement learning, to define intrinsic rewards~\citep{burda2018exploration}, or for planning and control~\citep{wahlstrom2015icmlw,finn2017icra,watter2015neurips}.

Recently, large-scale adversarial generative modeling of images has led to highly realistic generations, even at high resolutions~\citep{karras2017progressive,brock2018large,karras2018style}. Modeling of video~\citep{tulyakov2018mocogan, saito2018tganv2} has also seen impressive advances. \amended{\citet{clark2019dvdgan}} showed strong results on class-conditional generation of Kinetics-600, but the setting of video prediction was found to be surprisingly difficult in comparison. 
This may be due to the mode dropping behavior of GANs. Indeed, generating a video from scratch allows the network to learn just a few plausible types of sequences instead of needing to infer a plausible continuation for any sequence. 

In this work we focus on improving DVD-GAN-FP~\cite{clark2019dvdgan} in two ways. 
First, following \amended{the authors' }observation that the architecture of the discriminator is key for efficient training on large-scale datasets, we propose alternative decompositions for the discriminator and conduct an empirical study, validating a new architecture that achieves faster convergence and \amended{yields improved video quality.}

Second, we draw inspiration from recent state-of-the-art approaches for video prediction~\citep{hao2018cvpr,gao2019cvpr} 
that predict the parameters of transformations, \amended{use them} to warp the conditioning frames, and combine \amended{the result} with direct generation.
On the one hand, transformation-based prediction can use information in the conditioning frames, \amended{bypassing the need to learn to decompress representations of the relevant elements.}
On the other, in the presence of dis-occlusion, of appearing objects, or of new regions of the scene unfolding due to camera motion, warping past information seems an overly complex task in comparison with generating these pixel values directly.
The two approaches are hence highly complementary. 
We seek to incorporate such transformations in the DVD-GAN-FP architecture, to provide sufficient flexibility to model camera and object motion in a straightforward fashion, while retaining scalability of the overall architecture. 
For this purpose, we introduce the \gls{tsru}, a novel recurrent unit
that predicts motion-like features, and uses them to warp its past hidden state.
It then refines the obtained predictions by generating features to account for newly visible areas of the scene, and finally fuses the two streams.

To summarize, our contributions are as follows:
\begin{itemize}[leftmargin=*]
\item We conduct a systematic performance study of discriminator decompositions for DVD-GAN-FP and propose an architecture that achieves faster convergence and yields higher performance than previous approaches.

\item We propose a novel transformation-based recurrent unit to make the generator more expressive. This module is \amended{better suited for large-scale training} than existing transformation-based alternatives, and we show that it brings substantial performance improvements over classical recurrent units.

\item Our final approach, combining these two improvements, leads to a large improvement on the Kinetics-600 video prediction benchmark. 
Qualitatively, our model yields highly realistic frames and motion patterns. 

\end{itemize}

\section{Background and related work}
\amended{
\subsection{Video generation and GANs}
Since the introduction of the next frame prediction task~\citep{ranzato14video,srivastava15unsupervised}, video prediction and generation have become increasingly popular research topics. An important line of work has focused on extending ideas from generative modeling of images to these tasks~\citep{mathieu2015deep, vondrick2016generating, xue2016neurips, babaeizadeh2017stochastic, denton2018stochastic, kalchbrenner2017video}.

Generative Adversarial Networks (GANs)~\citep{goodfellow2014generative} define a minimax game between a \textit{discriminator} $\disc$ learning to distinguish real data from generated samples, and a \textit{generator} $\gen$ learning to minimize the likelihood of the discriminator classifying generated samples as generated.
In this work, we base our network architecture off DVD-GAN-FP~\citep{clark2019dvdgan}, which has built on the design principles combined in BigGAN for stable, large-scale training of image generative adversarial networks~\cite{brock2018large,miyato2018spectral,miyato2018cgans,lim2017geometric,dumoulin2017learned}.
The generator of DVD-GAN-FP is primarily a 2D convolutional residual network
with convolutional recurrent units interspersed between blocks at multiple resolutions to model consistency between frames.
The discriminator is composed of a spatial discriminator, assessing individual frames, and a spatio-temporal discriminator, assessing the concatenation of the conditioning and generated frames.
We provide an overview of DVD-GAN-FP in the appendix.

Many GAN approaches decompose the discriminator into multiple subnetworks. This includes multi-resolution approaches used for images~\citep{lapgan,stackgan,munit}, speech~\citep{binkowski2020iclr,kumar2019melgan} and video~\citep{tulyakov2018mocogan,clark2019dvdgan}. Some of these methods decompose the discriminator in a way which requires the existence of multiple generators, potentially sharing some hidden layers~\citep{saito2018tganv2, karras2017progressive}. While an important research direction, for the purposes of our analysis of discriminators, we do not consider such variants.

Progress on the task of video prediction has spanned multiple directions.
A number of works advocate for disentangling the factors of variation in the representations used by the models through careful design of the loss and network architecture: for example, between motion and content~\cite{tulyakov2018mocogan, villegas2017decomposing}, pose and appearance~\cite{denton2017unsupervised} or foreground and background~\cite{vondrick2016generating}. Object-centric representations have also been proposed~\cite{kosiorek2018neurips, zhu2018neurips}.
Some focus on alleviating error accumulation via architectural improvements~\cite{oliu2018eccv,villegas2017learning,vondrick2017cvpr}.
Several works also motivate the use of various reconstruction, ranking and perceptual losses~\citep{mathieu2015deep, jang2018icml, liu2018cvpr, li2018eccv, xiong2018cvpr}. 
Finally, another line of work reformulates video prediction in other feature spaces than the pixel space, such as high level image features~\citep{srivastava15unsupervised,vondrick2016anticipating}, optical flow~\citep{walker2015iccv}, dense trajectories~\citep{walker2016eccv} and segmentation maps~\citep{luc2017iccv, luc2018eccv}.~\citet{jayaraman2019iclr} operate in the pixel space, but choose not to commit to a fixed time offset between the conditioning frames and the prediction.

}

\subsection{Kernel-based and vector-based transformations}
\label{sec:warping-approaches}
 
Tranformation-based models for video prediction rely on differentiable warping functions to generate future frames as a function of past frames.
Following the terminology introduced by~\citet{reda2018eccv}, they fall into one of two families of methods: \emph{vector-based} or \emph{kernel-based}. 

Vector-based approaches consist in predicting the coordinates of a grid used to differentiably sample from the input~\cite{jaderberg2015neurips}.
This kind of approach has been used in conjunction with optical flow estimation~\citep{ranzato14video,patraucean2016iclrw, reda2018eccv}.
\citet{liu2017iccv} extend this idea, predicting a \emph{3D pseudo-optical flow field} used to sample from the entire sequence of conditioning frames \amended{via} tri-linear interpolation. 

Kernel-based approaches predict the parameters of dynamic, depth-wise locally connected layers~\citep{van2017transformation,xue2016neurips, finn2016unsupervised, vondrick2017cvpr, reda2018eccv}.
These predicted parameters can be shared across spatial locations, forcing each spatial position to undergo the same transformation.
To allow for varying motion across locations,~\citet{xue2016neurips} use several such transformations, and combine them using a predicted vector of weights at each position, to yield per-pixel filters.
Predicting a fixed number of depth-wise transformation kernels alongside a per-spatial-location weighting over transformations can be seen as a factorized version of predicting a full depth-wise transformation at each position.
\amended{
We therefore refer to the two approaches as \emph{factorized} and \emph{pixelwise}.}
\citet{finn2016unsupervised} explore both and find them to perform similarly.
\amended{
We refer to the appendix for a more formal description and an illustration for each.}

\subsection{Spatial recurrent units for video processing}

\label{sec:rec-unit-approaches}

\citet{shi2015neurips} and~\citet{ballas2015delving} introduce convolutional variants of the \gls{lstm}~\citep{hochreiter1997lstm} and \gls{gru} modules~\citep{cho2014emnlp}.
Our proposed recurrent unit builds on the convolutional recurrent units employed by DVD-GAN-FP, \Glspl{convgru}, which produce output $h_t$ based on input $x_t$ and previous output $h_{t-1}$ as:

\vspace{-0.5em}

\begin{align}
    \centering
    &r = \sigma(W_{r} \star_k [h_{t-1}; x_t] + b_r) \label{eq:convgru-r} \\
    &h^{\prime}_t = r \odot h_{t-1} \label{eq:convgru-hp} \\
    &c = \rho(W_{c} \star_k [h^{\prime}_t ; x_t] + b_c) \label{eq:convgru-c} \\
    &u = \sigma(W_{u} \star_k [h_{t-1}; x_t] + b_u) \label{eq:convgru-u} \\
    &h_t = u \odot h_{t-1} + (1 - u) \odot c \label{eq:convgru-h}
\end{align}

Here $\sigma$ is the sigmoid function, $\rho$ is the chosen activation function, $\star_{k}$ represents a $k \times k$ convolution, and the $\odot$ operator represents elementwise multiplication.

With similar motivations to ours,  
\citet{xu2018cvpr} propose a recurrent unit for structured video prediction,
relying on dynamic prediction of the weights used in the \gls{convlstm} update equations.
To avoid over-parameterization, the authors propose a shared learned mapping of feature channel differences to corresponding $2D$ kernels. 
This requires processing $\mathcal{O}(C^2)$ inputs\amended{, where $C$ is the number of channels of the hidden state,} which is prohibitively expensive in our large-scale setting.
We refer the interested reader to the appendix for a more detailed discussion. 

\citet{shi2017neurips} introduce \trajgru, which also warps the past hidden state to account for motion\amended{, using vector-based transformations}, and provides the result as input to all gates of a \gls{convgru}. 
\amended{
Vector-based transformations allow modeling of arbitrarily large motions with a fixed number of parameters,
but they require random memory access, whereas modern accelerators are better suited for computation formulated in terms of matrix multiplications.
As a result, in our work, we employ kernel-based transformations.}
We first propose a straightforward kernel-based extension of TrajGRU, called K-\trajgru.
We then formulate a novel transformation-based recurrent unit that is simpler and more directly interpretable.

\section{Large-scale transformation-based video prediction}
\label{s:dvdgan}

\begin{figure*}
\centering
\includegraphics[width=\textwidth]{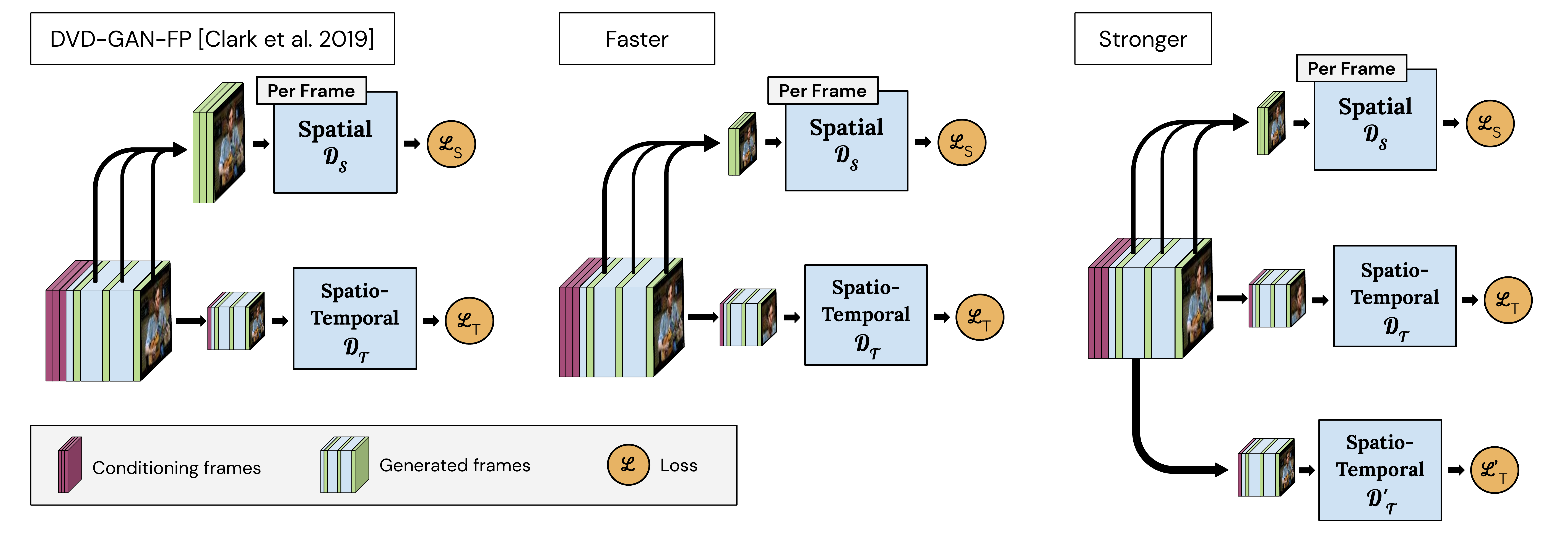}
\caption{
A visualization of the inputs each discriminator must judge. From left to right: 
the original DVD-GAN-FP discriminator decomposition processes full resolution frames and downsampled videos;
our proposed \emph{faster} decomposition processes downsampled frames and cropped videos; our proposed \emph{stronger} decomposition additionally processes downsampled videos.
}
\label{fig:diagram}
\end{figure*}

\subsection{Discriminator decompositions}
\label{ss:dd}

Architectures that can effectively leverage a large dataset like Kinetics-600 require careful design.
While the generator must create temporally consistent sequences of plausible frames, the discriminator must also be able to assess both temporal consistency and overall plausibility.

Both~\citet{tulyakov2018mocogan} and DVD-GAN-FP~\citep{clark2019dvdgan} employ two discriminators: a \textit{Spatial Discriminator} $\disc_S$ and a \textit{Temporal Discriminator} $\disc_T$, which respectively process individual frames and video clips.
While~\citet{tulyakov2018mocogan} motivate the use of $\disc_S$ mainly by improved convergence, DVD-GAN-FP makes the two discriminators complementary in order to additionally decrease computational complexity. 
In DVD-GAN-FP, $\disc_S$ assesses single frame content and structure by randomly sampling $K$ full-resolution frames and judging them individually, while $\disc_T$ is charged with assessing global temporal structure by processing videos \amended{spatially downsampled by a factor $s$.}

\amended{To further improve efficiency,} we propose an alternative split of the roles of the discriminators, with $\disc_S$ judging per-frame global structure, while $\disc_T$ critiques local spatio-temporal structure.
We achieve this by downsampling the $K$ randomly sampled frames fed to $\disc_S$ by a factor $s$, and cropping $T \times H/s \times W/s$ clips inside the high resolution video fed to $\disc_T$, where $T$, $H$, $W$ correspond to time, height, and width dimension of the input.
\amended{Compared with DVD-GAN-FP,} this reduces the number of pixels to process per video, from $K \times H \times W + T \times H/s \times W/s$ to $(K+T) \times H/s \times W/s$. 
We call this decomposition DVD-GAN-FP$_{faster}$.
In practice, we follow DVD-GAN-FP and choose $s=2$.

The potential blind spot of our new decomposition is global spatio-temporal coherence. For instance, if the motion of two spatially distant objects is correlated, then this decomposition will fail to teach $\gen$ that one object must move based on the movement of the other. 
To address this limitation, we investigate a second configuration which consists in combining our proposed decomposition with a second temporal discriminator $\disc^{'}_T$ assessing the quality of the downsampled generated videos, like in DVD-GAN-FP, and call this decomposition DVD-GAN-FP$_{stronger}$.
We summarize these configurations in Figure~\ref{fig:diagram} and Table~\ref{tab:comparison-discriminator-decompositions}.

\subsection{Transformation-based Recurrent Units}

\subsubsection{Stacked kernel-based warping for efficient, multi-scale modeling of motion}

\amended{As motivated in Section~\ref{sec:rec-unit-approaches}, we favour kernel-based transformations over vector-based ones. We}
rely on the stacking of multiple such transformations in the feature hierarchy of the generator to allow the modeling of large motion in a multi-scale manner, while keeping the number of parameters under control.

We first study a kernel-based extension of \trajgru, whose equations we report in the appendix. 
Inspired by recent work that leverages complementary pixel-based generations and transformation-based predictions~\citep{hao2018cvpr,gao2019cvpr}, we also propose a novel recurrent unit whose design is simpler and more directly interpretable, which we describe next.

\subsubsection{TSRU}

\amended{To combine transformation-based prediction with direct generation,} we begin by observing that the final equation of \gls{convgru}, Eq~(\ref{eq:convgru-h}), already combines past information $h_{t-1}$ with newly generated content $c$.
Similarly, our recurrent unit combines $\tilde{h}_{t-1}$, obtained by warping $h_{t-1}$, with newly generated content $c$.
Instead of computing the reset gate $r$ and resetting $h_{t-1}$, we compute the parameters of a transformation $\theta$, which we use to warp $h_{t-1}$. The rest of our model is unchanged (with $\tilde{h}_{t-1}$ playing the role of $h^{\prime}_t$ in $c$'s update equation, Eq~(\ref{eq:convgru-c})). \amended{Finally, our} module is described by the following equations:

\vspace{-1.5em}

\begin{align}
    \centering
    &\theta_{h, x} = f(h_{t-1}, x_t) \label{eq:tsru-theta}\\
    &\tilde{h}_{t-1} = w(h_{t-1} ; \theta_{h, x}) \\
    &c = \rho(W_c  \star_k [ \tilde{h}_{t-1} ; x_t] + b_c) \label{eq:tsru-c}\\
    &u = \sigma (W_u \star_k [h_{t-1}; x_t] + b_u) \label{eq:tsru-u}\\
    &h_t = u \odot \tilde{h}_{t-1} + (1 - u) \odot c \label{eq:tsru-h}, 
\end{align}

\begin{figure*}
\centering
\includegraphics[width=0.9\textwidth]{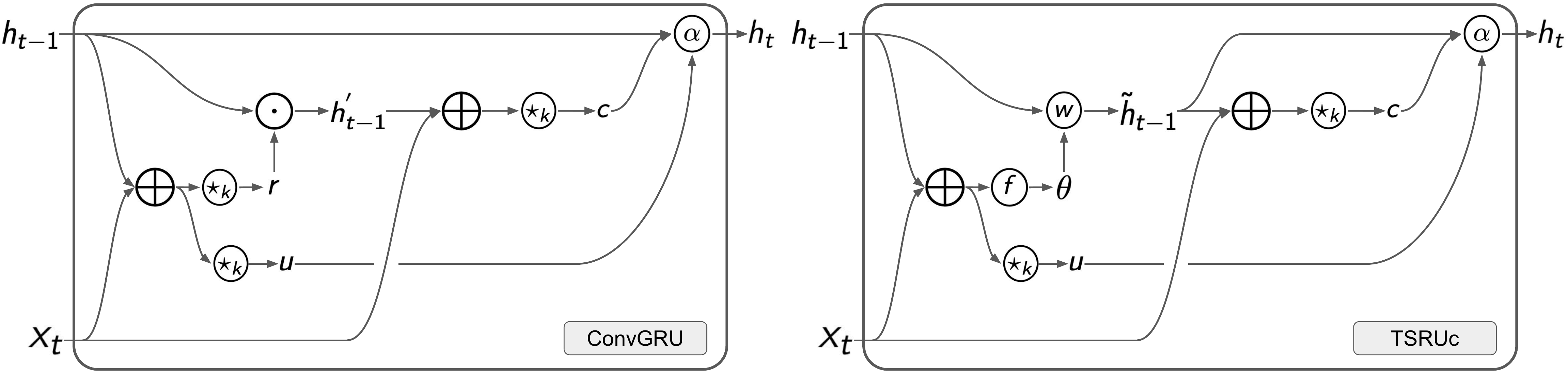}
\caption{Information flow for \gls{convgru} (left) and our proposed \tsru{}$_c$ (right); where $\alpha$ represents elementwise convex combination with coefficient provided by $u$; $\star_k$, convolution with kernel size $k$; $\bigoplus$, concatenation and $\bigodot$, elementwise multiplication. Other notations follow from Eqs (\ref{eq:tsru-theta})-(\ref{eq:tsru-h}).}
\label{fig:tsru}
\end{figure*}

\vspace{-0.5em}

where $f$ is a shallow convolutional neural network, $w$ is a warping function described next, and the rest of the notations follow from Eqs~(\ref{eq:convgru-r})-(\ref{eq:convgru-h}).
We have designed this recurrent unit to closely follow the design of \gls{convgru} and call it \tsru{}$_c$. 
\amended{We provide an illustration in Figure~\ref{fig:tsru}.}
We study two alternatives for the information flow between gates. \tsru{}$_p$ computes $\theta$, $u$ and $c$ in parallel given $x_t$ and $h_{t-1}$, yielding the following replacement for Eq~(\ref{eq:tsru-c}): $c = \rho(W_c  \star_k [ h_{t-1} ; x_t] + b_c)$.

At the other end, \tsru{}$_s$ computes each intermediate output in a fully sequential manner:
like in \tsru{}$_c$, $c$ is given access to $\tilde{h}_{t-1}$, but additionally, $u$ is given access to both outputs $\tilde{h}_{t-1}$ and $c$, so as to make an informed decision prior to mixing. This yields the following replacement for Eq~(\ref{eq:tsru-u}): $u = \sigma (W_u \star_k [\tilde{h}_{t-1}; c] + b_u)$. 
\amended{We also illustrate these variants in the appendix.}

The resulting modules are generic and can be used as drop-in replacements for other recurrent units, which are already widely used for generative modeling of video.
Provided that the conditioning frame encodings are used as initialization of its hidden representation, as in the DVD-GAN-FP architecture, this unit can predict and account for motion in the use it makes of past information, if this is beneficial.
When interleaved at different levels of the feature hierarchy in the generator, like in DVD-GAN-FP, motion can naturally be modeled in a multi-scale approach.
The modules are designed in such a way that a pixel-level motion representation can emerge, but we do not enforce this. In particular, we do not provide any supervision (e.g. ground truth flow fields) for the predicted motion-like features, instead allowing the network to learn representations in a data-driven way.

We study both factorized and pixelwise kernel-based warping (\emph{c.f.} Section~\ref{sec:warping-approaches} \amended{and appendix}).
Specifically, for factorized warping, we predict a sample-dependent set $\mathbf{w}$ of $N$ $2D$-convolution kernels of size $k \times k$ and a selection map $\mathbf{S}$, where each spatial position holds a vector of probabilities over the $N$ warpings, which we use as coefficients to linearly combine the basis weight vectors into per-pixel weights.
This map is intended to act as a pseudo-motion segmentation map, splitting the feature space into regions of homogeneous motion.
In practice, we choose $N=k^2$ and $k=3$.
Pixelwise kernel-based warping consists in predicting a $k\times k$ weight vector for each spatial position, which is used locally to weigh the surrounding values of the input in a weighted average. 
We denote this approach by \gls{ptsru}.

\section{Experiments and Analysis} 

\subsection{Experimental set up}
\label{ss:setup}

\paragraph{Datasets} Kinetics is a large dataset of YouTube videos intended for action recognition tasks. There are three versions of the dataset with increasing numbers of classes and samples: Kinetics-400~\citep{kay2017kinetics}, Kinetics-600~\citep{carreira2018short} and Kinetics-700~\citep{carreira2019short}. 
Here we use the second, Kinetics-600, to enable fair comparison to prior work. This dataset contains around 500,000 10-second videos at 25 FPS spread across 600 classes. Kinetics is often used in the action recognition field, but has only recently been used for generative purposes.~\citet{li2018video} and~\citet{balaji2018tfgan} used filtered and processed versions of the dataset, and more recently the entire unfiltered dataset has been used for generative modeling with GANs as well as autoregressive models~\citep{clark2019dvdgan, weissenborn2019scaling}. Following these works, we resize all videos to $64 \times 64$ resolution and train on randomly-selected 16 frame sequences from the training set.
Specifically, $\gen$ is trained to predict 11 frames conditioned on 5 input frames.
Unfortunately, the Kinetics dataset in general is not covered by a license which would allow for showing random data samples in a paper. For this reason, for all qualitative analysis and shown samples in this work, we use conditioning frames taken from the UCF-101 dataset~\citep{soomro2012ucf101}, though  the underlying models are trained completely on Kinetics-600. \amended{Finally, we extend our analysis on UCF-101~\cite{soomro2012ucf101} and BAIR Robot Pushing dataset~\cite{ebert2017self}, presented in the appendix.}

\paragraph{Metrics} Traditional measures for video prediction models such as Peak Signal-to-Noise Ratio (PSNR) and Structural Similarity Index Measure (SSIM)~\citep{wang2004image} rely on the existence of ground truth continuations of the conditioning frames and judge the generated frames by their similarity to the real ones. While this works for small time scales with near deterministic futures, these metrics fail when there are many potential future outcomes.~\citet{unterthiner2018towards} propose the Fr\'echet Video Distance (FVD), an adaption of the Fr\'echet Inception Distance (FID) metric~\citep{heusel2017gans} which judges a generative model by comparing summary statistics of a set of samples with a set of real examples in a relevant embedding space. For FVD, the logits of an I3D network trained on Kinetics-400~\citep{carreira2017quo} are used as the embedding space. We adopt this metric for ease of comparison with prior work. We also report the Inception Score (IS).

\paragraph{Training details}
All models were trained on 32 to 512 TPUv3 accelerators using the Adam~\citep{kingma2014adam} optimizer. 
In all our experiments, we validate the best model according to performance on the training set.
Like BigGAN~\citep{brock2018large}, we find it important to use cross-replica batch statistics where the per-batch mean and variance are calculated across all accelerators at each training step. More details on our training setup are provided in the appendix. \amended{We will make the code\footnote{Our training script is heavily tied to proprietary code, and there is no simple way to release it. However, upon acceptance, we will release functions relating to data processing, models, losses and metrics.} available, as well as models trained on UCF-101 and BAIR.}

\subsection{Comparing Discriminator Decompositions}
\label{sec:discriminator-decomposition-results}
\begin{table*}[t]
\caption{A comparison of different discriminator decompositions for video prediction trained for a fixed time budget evaluated on the Kinetics 600 validation set. We describe each in terms of the types of sub-discriminators it contains: frame-level (F), downsampled frame-level (dF), clipped video (clV), downsampled video (dV) and cropped video (crV).
Average step duration in milliseconds (ms) and maximum memory consumption (GB) are measured for a single forward-backward training step, averaged over 30 seconds.
We report the step budget for each run in number of thousands of steps~(K).
}
\label{tab:comparison-discriminator-decompositions}
\vskip 0.0in
\begin{center}
\begin{small}
\begin{sc}
\begin{tabular}{l|ccccc|cc|c|cc}
\toprule
            & F         & dF        & clV       & dV        & crV 
            & IS            & FVD
            & K & ms        & GB    \\
\midrule
Clip        &           &           & \checkmark&           &
            & 10.9          & 59.6 
            & 758 & 488               & 3.49                    \\ Downsampled &           &           &           & \checkmark&
            & 8.9              & 122.1
            & 1000 & 370               &  3.01                     \\ 

Cropped     &              &            &           &           & \checkmark
            & 9.4          &  159.9
            & 997 & 371               &  3.10                     \\ Cropped + Downsampled     &           &           &           & \checkmark  & \checkmark
            & 11.5              & 44.1
            & 792 & 467               & 3.30                  \\ \midrule
MoCoGAN-like $K=1$& \checkmark&           & \checkmark&           &
            & 11.3             & 47.4
            & 651 & 568               & 3.95                \\ MoCoGAN-like $K=8$& \checkmark&           & \checkmark&           &
            & 11.4              & 46.2
            & 561 & 660               & 3.95          \\ DVD-GAN-FP (ours)      & \checkmark&           &           & \checkmark&
            & 10.9              & 55.0
            & 689 & 537               & 3.53             \\ \midrule
DVD-GAN-FP$_{faster}$&        & \checkmark&           &           & \checkmark
            & 11.7              & 46.1
            & 817 & 453               & 3.54             \\ DVD-GAN-FP$_{stronger}$ &        & \checkmark&           & \checkmark& \checkmark
            & \textbf{11.8}     & \textbf{39.3}
            & 675 & 548               & 3.65               \\ \bottomrule
\end{tabular}
\end{sc}
\end{small}
\end{center}
\end{table*}
\amended{
We begin by comparing several discriminator decompositions, including a MoCoGAN-like decomposition~\cite{tulyakov2018mocogan}, DVD-GAN-FP~\cite{clark2019dvdgan}, and our proposed decompositions,
on the Kinetics video prediction benchmark~\cite{weissenborn2019scaling}.
We fix the number of frames $K$ sampled for the spatial discriminator to $8$.
For \mocogan{}, we also evaluate the authors' original setting, where $K=1$.}
Finally, for each of the discriminator decompositions, we evaluate a baseline where only the temporal discriminator is used, denoted respectively by \emph{Clip} (for \mocogan{}), \emph{Downsampled} (for DVD-GAN-FP), \emph{Cropped} (for DVD-GAN-FP$_{faster}$) and \emph{Cropped \& Downsampled} (for DVD-GAN-FP$_{stronger}$).

Because we are interested in architectures which improve real experimentation time, we fix the step budget for each method according to its average step duration, measured over $30$ seconds for a batch size of $512$, and make sure that we do not account for any lag introduced by data loading or preprocessing.
For these normalized step budgets, training takes approximately 103 hours, corresponding to $1M$ iterations for the fastest one. 
We summarize step budgets and results in Table~\ref{tab:comparison-discriminator-decompositions}.

For all four discriminator decompositions, the addition of a frame discriminator yields large improvements in FVD, along with improving convergence speed as previously noted by~\citet{tulyakov2018mocogan}.
We also observe that the \emph{Downsampled} and \emph{Cropped} baselines obtain poor performance compared to the other approaches, consistent with our observation that the \emph{Downsampled} baseline can only assess global spatio-temporal structure, and the \emph{Cropped} baseline can only assess local structure. In contrast, a combination of the two (\emph{Cropped + Downsampled}) largely improves on the \emph{Clip} baseline, which has access to all conditioning and predicted frames, and obtains improved performance with respect to the two \mocogan{} variants, while being $29\%$ faster to train.
Next, we find that in this setting, DVD-GAN-FP yields an improvement in terms of average step duration over the \mocogan{} baselines, but lags behind in terms of prediction quality.
As expected, the alternative decomposition we propose has a much shorter average step duration than the \mocogan{} and DVD-GAN-FP baselines, and additionally closes the performance gap with \mocogan{}.
Adding a spatial discriminator to the \emph{Cropped \& Downsampled} baseline yields our final proposed decomposition, DVD-GAN-FP$_{stronger}$, which obtains a substantial improvement over previous approaches. 

\subsection{Comparing Recurrent Units}
\label{sec:recurrent-units-results}

We now report a comparison between recurrent units in Table~\ref{tab:comparison-recurrent-units}. Here, we use the fastest of our discriminator decompositions, DVD-GAN-FP$_{faster}$, so as to maximize the number of training steps which can be afforded for each configuration in this large-scale setting. We train all models for 500k steps.
Our setup is otherwise identical to the one presented in Section~\ref{sec:discriminator-decomposition-results}.
We find that all our proposed transformation-based approaches perform better than using \gls{convgru} or \gls{convlstm}, suggesting that the increased expressivity of the generator is beneficial for generation quality, with the best results observed for \gls{tsru}$_p$ in this setting.
Interestingly, we observe that \gls{ptsru} performs slightly worse than its \gls{tsru}$_s$ counterpart (which is the least competitive of our \gls{tsru} designs); 
suggesting that the factorized version may be more effective at predicting structured, globally consistent motion-like features.
As a result, we do not investigate other versions of the \gls{ptsru} design.
Finally, we observe a small performance improvement for all \gls{tsru} compared with K-\trajgru{}, suggesting that its simplified design facilitates learning. 

\begin{table}[t]
\parbox[t][][t]{.41\linewidth}{
\caption{Comparison of recurrent unit designs for video prediction on Kinetics 600 validation set.
All models are trained for 500k steps.
}
\label{tab:comparison-recurrent-units}
\begin{center}
\begin{small}
\begin{sc}
\begin{tabular}{lccc}
\toprule
      & IS ($\uparrow$) & FVD ($\downarrow$) \\
\midrule
ConvLSTM &  11.4         & 51.3          \\ ConvGRU & 11.4          & 49.4          \\ 

\midrule
K-TrajGRU
            &  11.6         & 45.9                  \\ TSRU$_c$      & 11.6        & 44.2          \\ TSRU$_p$      & 11.5        & $\mathbf{43.7}$          \\ TSRU$_s$      & 11.6        & 45.8          \\ PTSRU$_s$       & 11.5      & 47.1          \\ \bottomrule
\end{tabular}
\end{sc}
\end{small}
\end{center}
}
\hfill
\parbox[t][][t]{.55\linewidth}{
\caption{Video prediction performance on Kinetics-600 test set, without frame skipping. FVD is computed using test set video statistics. We predict 11 frames conditioned on 5 input frames.
We provide unbiased estimates of standard deviation over 10 runs.
}
\label{tab:soa-video-prediction-kinetics600}
\vskip 0.15in
\begin{center}
\begin{small}
\begin{sc}
\begin{tabular}{lcccc}
\toprule
Method                  & IS ($\uparrow$)        & FVD ($\downarrow$) \\
\midrule
Video Transformer       &  --       & 170 $\pm$ 5    \\
DVD-GAN-FP     & -- & 69.15 $\pm$ 1.16 \\
\midrule
\dvdganpp{} & $12.54 \pm 0.06$ & $\mathbf{25.74 \pm 0.66} $ \\ 
\bottomrule
\end{tabular}
\end{sc}
\end{small}
\end{center}
}
\end{table}

\subsection{Scaling up}
\label{sec:large-scale-results}

Putting it all together, we combine our strongest decomposition DVD-GAN-FP$_{stronger}$ with TSRU$_p$. 
We call this model \dvdganpp{}, for ``Transformation-based \& TrIple Video Discriminator GAN''.
We increase the channel multiplier (described in the appendix) from 48 to 120 and train this model for 700,000 steps. 
We report the results of this model evaluated on the test set of Kinetics 600 in Table~\ref{tab:soa-video-prediction-kinetics600}, together with the results reported by~\citet{weissenborn2019scaling} and~\citet{clark2019dvdgan}.
We also provide unbiased estimates of standard deviation for 10 training runs. In comparison with the strong DVD-GAN-FP baseline, our contributions lead to a substantial
decrease of FVD. 

\begin{figure}[t]
    \centering
    \includegraphics[width=0.495\linewidth]{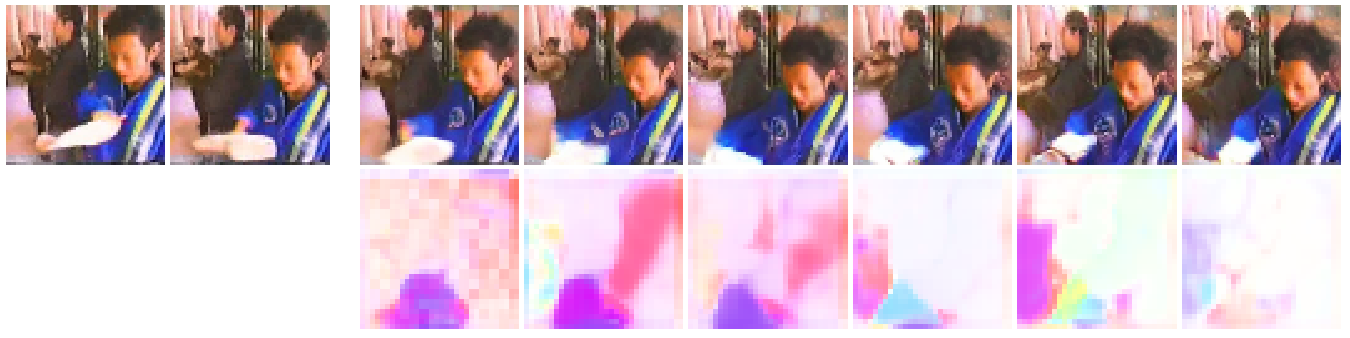} \hfill
    \includegraphics[width=0.495\linewidth]{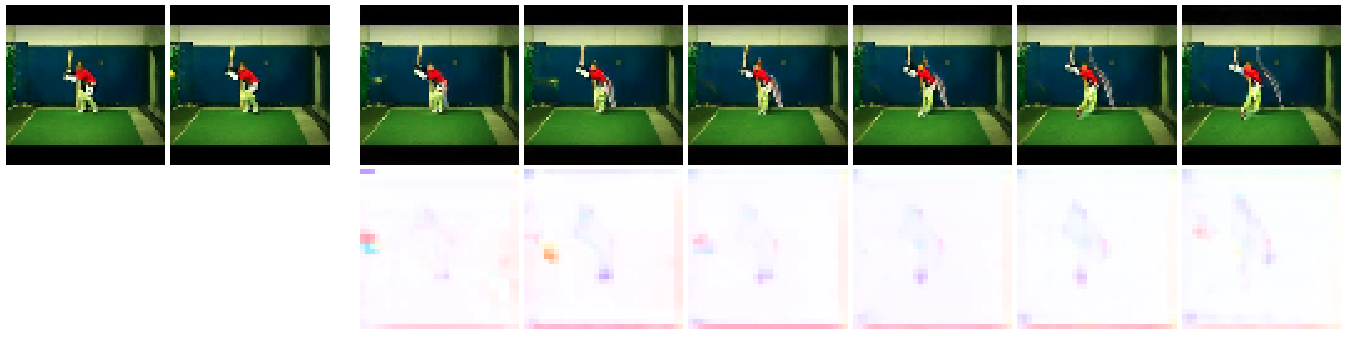} \\
    \includegraphics[width=0.495\linewidth]{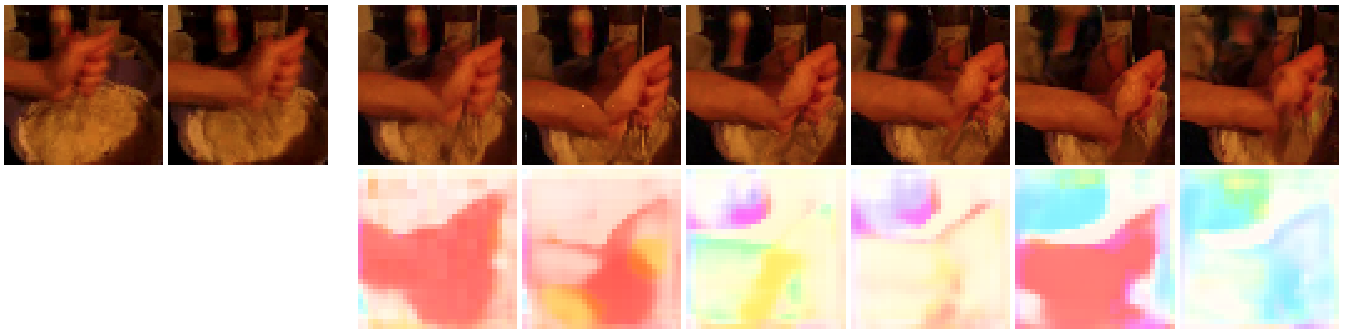} \hfill
    \includegraphics[width=0.495\linewidth]{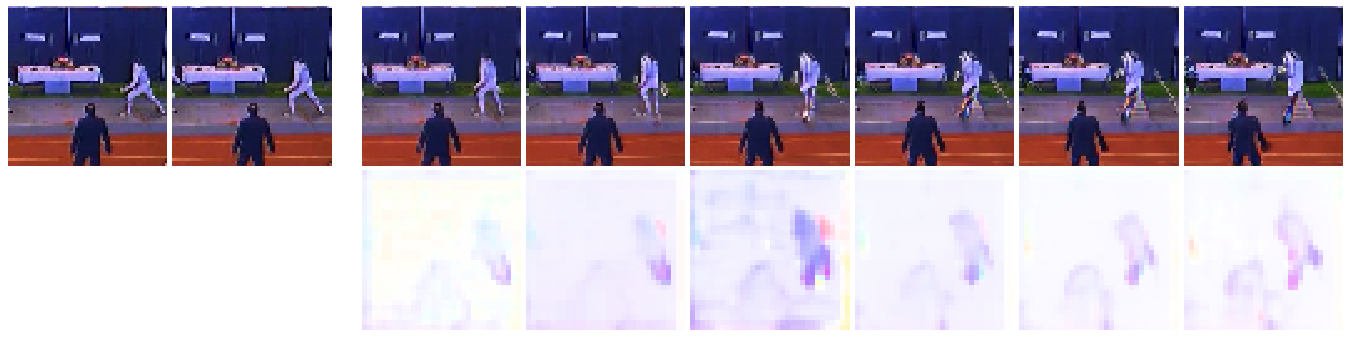} \\
    \caption{Qualitative evaluation on UCF-101 test set 1. 
    Top: predictions. Second row: motion field obtained by combining the two lower levels of predicted motion-like features. All sequences are temporally subsampled by two for visualization.}
    \label{fig:qualitative-motion-fields}
\hfill
\end{figure}

In Figure~\ref{fig:qualitative-motion-fields}, we show predictions and corresponding motion fields from our final model \dvdganpp{}.
We select them to demonstrate the interpretability of these motion fields for a number of samples.
We obtain them by interpreting the local predicted weights for a given spatial position as a probability distribution over flow vectors corresponding to each of the kernel grid positions and taking the resulting expected flow value.
They can be combined across resolutions by iteratively upsampling a coarse flow field, multiplying it by the upsampling factor to account for the higher resolution, and refining it with the next lower level flow field.
These predictions have been obtained on UCF-101 and also show strong transfer performance.
Though this is not entirely surprising given the similarity between the classes and content, it further emphasizes the generalization properties of our model. 
We provide \amended{an extended qualitative analysis including additional samples} in the appendix.

\amended{
Finally, we perform additional experiments on the BAIR Robot Pushing dataset, 
and on UCF-101, 
to demonstrate the generality of our method.
We further show that our model is able to generate diverse predictions on the BAIR dataset.
We report the results in the appendix.
}

\section{Conclusion}

Effectively training video prediction models on large datasets requires scalable and powerful network architectures. In this work we have systematically analyzed several approaches towards reducing the computational cost of GAN discriminators and proposed DVD-GAN-FP$_{faster}$ and DVD-GAN-FP$_{stronger}$, which improve wall-clock time and performance over the strong baseline DVD-GAN-FP. We have further motivated and proposed a family of transformation-based recurrent units -- drop-in replacements for standard recurrent units -- which further improve performance. Combining these contributions led to our final model, \dvdganpp{}, which obtains large performance improvements and is able to make diverse predictions.  
For future directions, we plan to investigate the use of these recurrent units for video processing tasks, as well as the usefulness of the representations that our models learn.
\medskip

\amended{
\section*{Broader Impact}

Video generative modeling is a well-established research problem, which can drive the development of general purpose technologies for modeling distributions of data arising from complex spatio-temporal phenomena. Kinetics-600 is a large, diverse and complex dataset, and as such, it is reasonable to expect that approaches that yield improvements on this dataset should generally prove useful in many other settings. Our proposed discriminator decomposition yields faster convergence and better performance. Our proposed transformation-based recurrent units provide additional flexibility to the generator, and are designed such that pixel-level motion representations can emerge in a data-driven way.  Such advances could be applied to the acceleration of climate models. Extensions of video prediction models could replace components for which physical counterparts are too computationally intensive or too uncertain. For example, high resolution cloud simulation~\citep{gentine2018grl} and prediction of short-term changes in the sea ice extent using satellite images have both been identified as a "high leverage" machine learning applications for tackling climate change~\cite{rolnick2019arxiv}. Our improvements could also generally benefit applications that involve decision-making in the real world, such as autonomous driving and robotics.

On a more pessimistic note, progress in the field of video generation models might be applied by malicious agents, to forge false, plausible elements in support of deceptive affirmations, thus contributing to the creation, confirmation and spread of beliefs. This can, in turn, have dramatic consequences. As the field progresses, methods that can detect generated images or videos automatically, such as~\citep{afchar2018mesonet,sabir2019recurrent,xuan2019generalization}, are therefore becoming increasingly important.~\citet{nguyen2019deep} provide a survey of such methods.

Using Kinetics as a benchmark might also inadvertently result in researchers overlooking privacy concerns, motivated by research incentives. Kinetics is a dynamic dataset, with a small number of videos occasionally removed, following their removal from YouTube~\cite{kay2017kinetics}. 
We believe that releasing frames or continuations for this dataset would therefore be unethical. The same concern applies to releasing generative models trained on this dataset, since we cannot guarantee that they will not be able to partially memorize some of the removed videos. To mitigate this risk, we hope that future research can continue to follow our evaluation procedure, when using this dataset, by presenting only aggregated metrics of performance, and by employing other datasets to demonstrate qualitative performance.

}

\section*{Acknowledgements}

We thank Jean-Baptiste Alayrac, Lucas Smaira, Jeff Donahue, Jacob Walker, Jordan Hoffman, Carl Doersch, Joao Carreira, Andy Brock, Yusuf Aytar, Matteusz Malinowski, Karel Lenc, Suman Ravuri, Jason Ramapuram, Viorica Patraucean, Simon Osindero and Chloe Rosenberg for their help, feedback and insights.

\bibliography{references}
\bibliographystyle{plainnat}

\appendix

\section{Overview of DVD-GAN-FP}
\label{ap:model}

Except for the improvements discussed in the main text, our model is identical to DVD-GAN-FP \citep{clark2019dvdgan}. This model consists of an encoder $\enc$, a generator $\gen$ and a discriminator $\disc$. We give an overview of the architecture of each component here, and refer the reader to the paper for additional details.
\amended{
Both BigGAN and DVD-GAN were initially formulated in a class-conditioned setting. 
For simplicity, DVD-GAN-FP replaces the class label with a dummy constant zero class label wherever a class label is used.

$\gen$ is primarily a 2D convolutional residual network based on the BigGAN architecture \citep{brock2018large}, with convolutional recurrent units interspersed between blocks at multiple resolutions to model consistency between frames. 
First, a feature vector is obtained by concatenating a latent sample $z$ drawn from a normal distribution and features linearly obtained from the class label, both of dimension $128$.
Spatial features of resolution $8\times8$ are obtained via a reshaped affine transformation of this feature vector.
These features are repeated along the time dimension, and passed for each time-step as input to the first convolutional recurrent unit. 
The features output at each time step then undergo framewise processing:
a convolutional layer, 
followed by two identical BigGAN G residual blocks, the second of which performs nearest-neighbour spatial upsampling by 2.
The recurrent unit and the framewise processing form the first stage of the architecture.
Another two similar stages are employed, leading to features of resolution $64\times64$.
At each stage, the recurrent unit's hidden state is initialized with the features extracted by the encoder at that resolution.
Finally, the predicted images are obtained through a final BatchNorm-ReLU, a convolutional layer and a tanh.
Batch normalization layers are also employed in each of the G residual blocks, independently for each time step, and conditioned on the 256-dimensional input feature vector containing the latent sample and the class information.
All linear and convolutional layers in the entire model have Spectral Normalization applied to their weights \cite{zhang2018self},
with the exception of the linear embedding mapping to the first spatial features.


All discriminators discussed in the main text are convolutional networks almost identical to BigGAN's discriminator, except in the input they take. 
Per-frame discriminators are entirely unchanged: they consist of five stages, each containing a single D residual block, the first four of which employ average pooling to downsample the features by 2. The final block is followed by a ReLU activation.
Features are then summed across the spatial dimensions, and passed to the projection head, further described in Section~\ref{ap-sec:loss}.
The spatio-temporal discriminators, which are responsible for assessing temporal consistency, process the inputs identically, with the exception that all convolutions in the first two residual blocks are replaced with 3D convolutions.

In order to condition $\gen$ on initial frames, an encoder $\enc$, identical to the spatial discriminator network, is applied to the conditioning frames.
For each RNN present in $\gen$, the features extracted by the encoder at the corresponding resolution are reshaped, by folding the time dimension inside the channel dimension.
The result is compressed using a single convolutional layer, to form the initial state for the corresponding RNN. %
In this way, all the recurrent units in $\gen$ are initialized with features that have knowledge of the conditioning frames.

We refer the reader to \cite{clark2019dvdgan} for the parametrization of the G and D residual blocks, and the corresponding hyperparameters, which our method leaves unchanged, unless stated otherwise.

}

\section{Kernel-based warping}

We give a more formal description of factorized and pixelwise kernel-based warping, as introduced in Section 2.3. Pixelwise kernel-based warping approaches employ a network $f$ to predict a set of parameters $\theta = \mathbf{W} \in \mathbf{R}^{H \times W \times k^2}$, given input $x$. 
These parameters are then used in depthwise, dynamic, locally-connected layers of kernel size $k$, taking as input $h \in \mathbf{R}^{H\times W \times C}$. 
In previous kernel-based video prediction approaches \citep{xue2016neurips,finn2016unsupervised,vondrick2017cvpr,reda2018eccv}, $h$ and $x$ are respectively the last conditioning frame and the concatenated conditioning frames. In the case of our recurrent units, $h$ represents the past hidden state $h_{t-1}$ and the input $x$ is formed by the concatenation of $h_{t-1}$ and the current input to the unit $x_t$.
Formally, at a given spatial position $(i,j)$, the $c^\text{th}$ dimension of the output vector $\tilde{h}_{i,j}$ is given by:
\begin{equation}
    \label{eq:warping}
    \tilde{h}_{i,j}[c] = \sum_{(m,n) \in \llbracket 0, k-1 \rrbracket^2} \mathbf{W}_{i,j}[mk +n] \cdot h_{i+m-(k-1)//2, j+n-(k-1)//2}[c],
\end{equation}

where $h$ has been padded to preserve spatial dimensions.

In the case of factorized kernel-based warping, $f$ predicts $\theta=(\mathbf{w}, \mathbf{S})$, where $\mathbf{w} \in \mathbf{R}^{k^2 \times N}$ and $\mathbf{S} \in \mathbf{R}^{H \times W \times N}$, which are combined to form a tensor $\mathbf{W} \in \mathbf{R}^{H \times W \times k^2}$ as follows. At $(i,j)$, the $q^\text{th}$ dimension of $\mathbf{W}_{i,j}$ is given by :

\begin{equation}
    \label{eq:w}
    \mathbf{W_{i,j}}[q] = \sum_{l=1}^{N} \mathbf{S}_{i,j}[l] \cdot \mathbf{w}[q, l].
\end{equation}

\begin{figure*}[h]
\centering
\includegraphics[width=\textwidth]{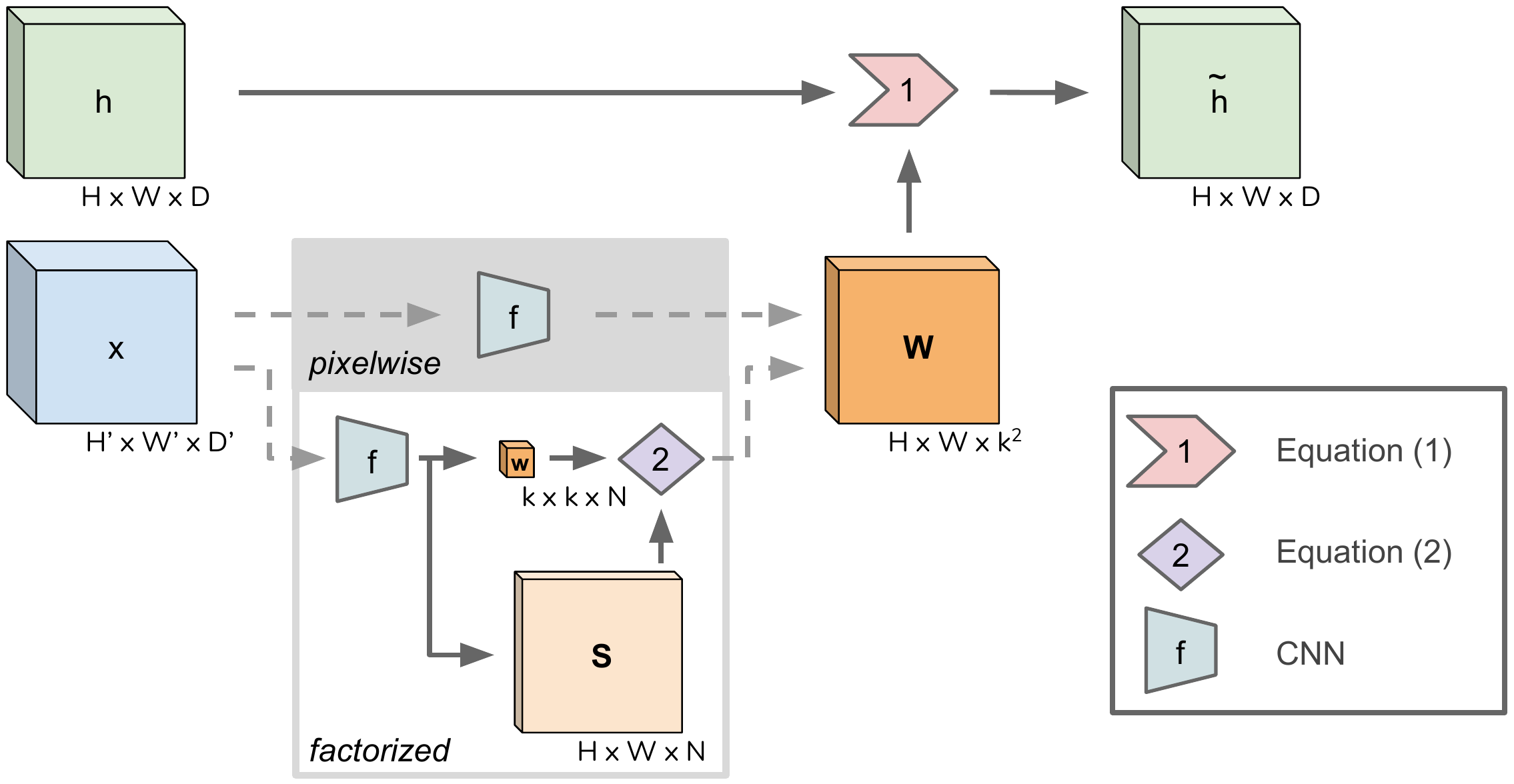}
\vspace{-1em}
\caption{\amended{Kernel-based warping. Given input $x$, a network $f$ predicts transformation parameters: a tensor of weights $\mathbf{W}$ for pixelwise warping, or a set of weights $\mathbf{w}$ and a map of coefficients $\mathbf{S}$ for factorized warping, which are then combined to obtain $\mathbf{W}$ (Equation~(\ref{eq:w})).
In both cases, the result $\mathbf{W}$ is used in a dynamic, locally connected, depth-wise layer, applied to $h$ (Equation~(\ref{eq:warping})).}}
\label{fig:warping}
\end{figure*}

$\mathbf{W}$ is then used as in the pixelwise warping case. We illustrate the procedure for pixelwise and factorized kernel-based warping in Figure~\ref{fig:warping}.

\amended{
\section{Additional experiments}

\subsection{BAIR Robot Pushing}

\label{ap-sec:bair-results}

The BAIR Robot Pushing dataset of robotic arm motion~\citep{ebert2017self} has been used as a benchmark in prior work on video prediction~\citep{babaeizadeh2017stochastic, denton2018stochastic,lee2018stochastic,unterthiner2018towards, clark2019dvdgan, weissenborn2019scaling}. 
The original BAIR Robot Pushing dataset contains 43,264 training samples and 256 test samples, each of 30 frames. 
\citet{unterthiner2018towards} have shown that the expected FVD between two non-overlapping 256-sized subsets of 16-frame videos randomly drawn from this dataset is on the order of $10^2$. 
This suggests that on such a small test set, this metric has a very important bias, impeding quantitative comparison to state-of-the-art results.
Instead, if we take two non-overlapping 30k-sized subsets, the expected FVD falls close to zero \citep{unterthiner2018towards}. 
We therefore wish to employ 30k samples to estimate the statistics of both ground truth and generated samples.
To do so, we propose to take a more balanced split of the data, using only the first $70\%$ of the train set for training, and the remaining $30\%$ augmented with the original 256 test samples for evaluation.

We train TrIVD-GAN-FP, using a channel multiplier in $\gen$ of 48, and in $\disc$ of 92, and learning rates of $2\cdot10^{-5}$ for $\gen$ and $3\cdot10^{-5}$ for $\disc$.
Like previous work, we condition on one frame and generate 15 future frames.
As on Kinetics, we validate models according to performance on the train set and report FVD on the test set for both splits.
We use the original split to compare to Video Transformer and DVD-GAN-FP, and our proposed balanced split, as a baseline for future work.

We report the obtained results on this dataset in Table~\ref{tab:bair-results}.
Despite substantial improvement over DVD-GAN-FP and Video Transformer~\cite{weissenborn2019scaling} on Kinetics, on BAIR,  \dvdganpp{} only sees slight improvement over DVD-GAN-FP, still higher than Video Transformer. Qualitatively, we do not observe a visual difference between the models, and echo comments from both papers that FVD may not be an adequate metric on BAIR, especially given the large bias of the metric resulting from the small sample size.
Given access to 30K samples to estimate the FVD with the more balanced split, our model reaches $31.8$ test FVD, estimated over $10$ evaluation runs.

\begin{table}
    \caption{\amended{FVD on BAIR Video Prediction, using the original (O) and balanced (B) splits of the BAIR dataset.
    On the original split, FVD is estimated using 256 samples, due to the small test set size, leading to a large bias for the FVD (on the order of $100$.) We propose a more balanced split of the data, to allow the use of 30K samples to estimate the test FVD.}}
    \label{tab:soa-video-prediction-bair}
    \vskip 0.15in
    \begin{center}
    \begin{small}
    \begin{sc}
    \begin{tabular}{lcccc}
    \toprule
    Method   & Split & FVD ($\downarrow$) \\
    \midrule
    SAVP & O &116.4    \\
    DVD-GAN-FP & O & 109.8     \\
    \dvdganpp{} & O & 103.3     \\
    Video Transformer & O & \textbf{94 $\pm$ 2}     \\
    \midrule
    \dvdganpp{} & B & \textbf{31.8 $\pm$ 0.2}     \\
    \bottomrule
    \label{tab:bair-results}
    \end{tabular}
    \end{sc}
    \end{small}
    \end{center}
    \vskip -0 in
\end{table}

\begin{figure}
    \centering
    \includegraphics[width=0.8\linewidth]{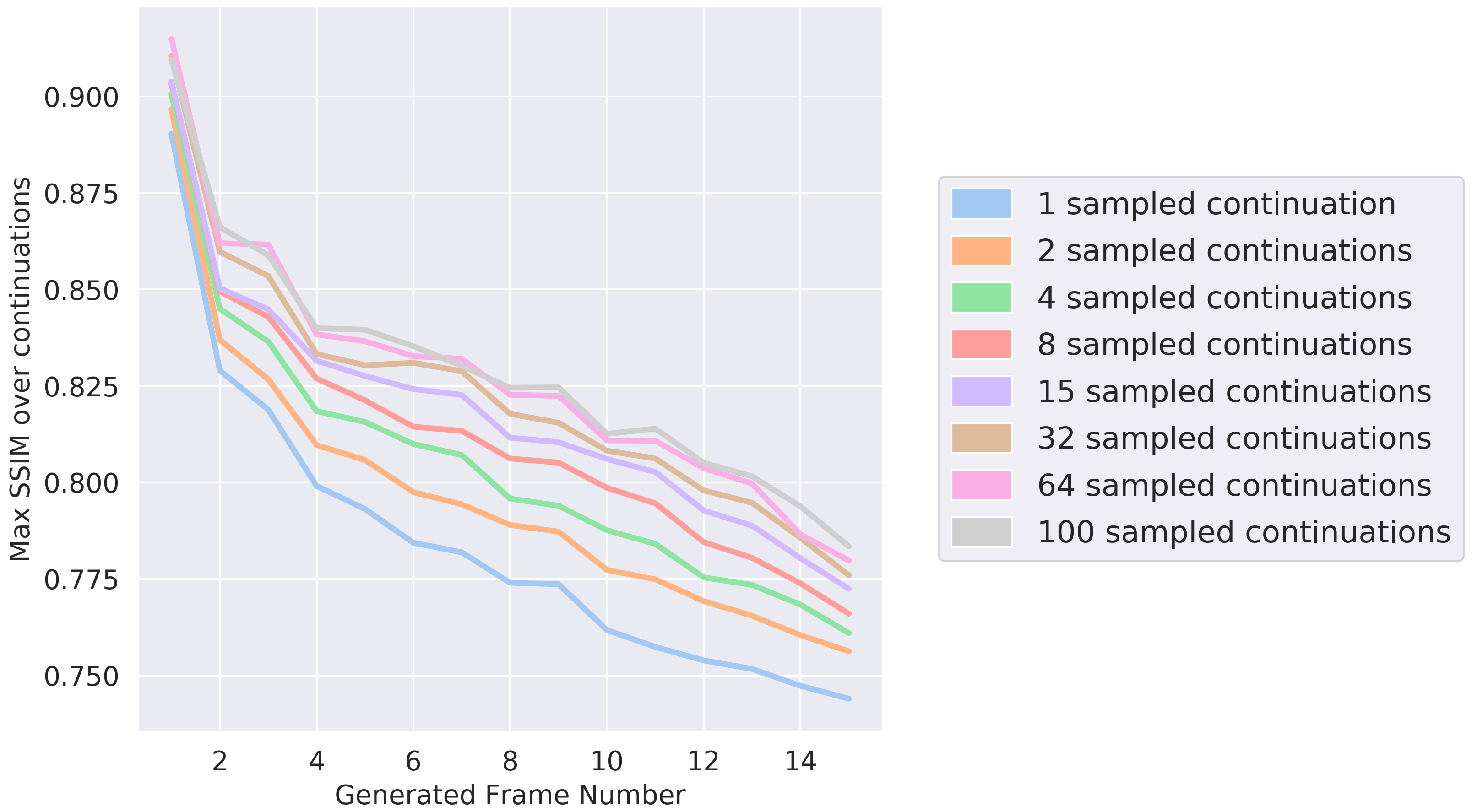}
    \vspace{-0em}
    \caption{
    Per-frame SSIM for \dvdganpp{} where SSIM is taken as the max over differing numbers of sampled continuations. Higher is better.}
    \label{fig:ssim}
\end{figure}

Finally, we report SSIM scores from BAIR, collected as in SAVP~\cite{lee2018stochastic}. This involves generating $\ell$ continuations of a single frame and selecting the one which maximizes the average frame SSIM, then reporting per-frame SSIMs for all ``best  continuations'' of conditioning frames. The results of this experiment are in Figure~\ref{fig:ssim}.
Notably, we see high correlation between SSIM and increasing $\ell$, meaning that generating more samples leads to a meaningfully diverse variety of futures, such that some are closer to the ground truth than others.

\subsection{UCF-101}

UCF-101~\cite{soomro2012ucf101} is dataset of 13,320 videos of human actions across 101 classes.
As a complement to the qualitative results we present, we also provide quantitative performance on this dataset. For the models with a channel multiplier of 48 trained for 1M iterations on Kinetics, TrIVD-GAN-FP improves upon DVD-GAN-FP from 117 test FVD to 95.8. 
The large model with a channel multiplier of 120 trained on Kinetics further improves this to 67.7. Finally, we train a model with channel multiplier of 92 on the UCF train set for 1M iterations, using the same hyperparameters as on Kinetics, and obtain our best results, of
$55.2\pm 0.64$  test FVD.
}

\section{Extended qualitative analysis}

We start by showing in Section~\ref{ap-sec:qual-comp} a qualitative comparison between recurrent units, using our medium-size models, described in Section~4.3 
of the main paper.
Next, in Sections~\ref{ap-sec:latent-repr} and \ref{ap-sec:success-failure}, we provide further qualitative analysis of the samples predicted by our final large-scale TrIVD-GAN-FP model, described in Section~4.4 
of the main paper.
To support our analysis, we provide accompanying videos at {\color{magenta}\url{https://drive.google.com/open?id=1fvmEm3gOWprWy2IuMFe4DuwEPahe9MwC}}.

All models have been trained on Kinetics-600 train set, and all predictions have been sampled randomly on the UCF-101 test set 1, due to restrictions on Kinetics-600.

\subsection{Qualitative comparison between ConvGRU and TSRU}
\label{ap-sec:qual-comp}

We now compare our proposed recurrent unit TSRU$_p$ with ConvGRU, using for each the best model obtained when trained for 1M steps.
We employ the truncation trick \citep{brock2018large}, as we find that this significantly improves both models' predictions.
Specifically, we employ moderate truncation of the latent representation, using threshold $0.8$.

We refer the reader to the accompanying video in folder \emph{C.1.},
showing a side-by-side comparison of randomly sampled predictions, with the ConvGRU baseline on the left, and the TSRU model on the right. We find that our proposed module yields a slight but perceptible improvement over ConvGRU, 
in that the model using TSRU is more often able to conserve a plausible object shape across frame predictions.

\subsection{Influence of the latent representation on the predictions}
\label{ap-sec:latent-repr}

Here, we provide a qualitative analysis of the influence of the latent representation on the predictions, using our large-scale TriVD-GAN-FP model.
In the directory \emph{C.2.}, we provide a comparison of samples obtained using, on the left constant zero latent representations $z$ (i.e. extreme truncation) and on the right random latent representations sampled from the same distribution as the one used during training (i.e. no truncation).

We find that in comparison with the constant zero latent representation samples, the non-truncated samples exhibit much more drastic changes in the predictions with respect to the input frames.
Camera motion, zooming or object motion tend to be more important in a number of samples when using the non-truncated distribution.
Effects like fade-to-black or novel object appearances, tend to appear regularly in the non-truncated examples, 
while we have not observed them in the samples that use a fixed zero latent representation. We point out examples of such effects in the annotated comparison (\emph{annotated$\_$constantz$\_$vs$\_$notruncation});
and also provide a non-annotated comparison (\emph{constantz$\_$vs$\_$notruncation}).
This points to a significant influence of the random variable on the predictions, as also evaluated quantitatively on the BAIR dataset in Section~\ref{ap-sec:bair-results}.

Finally, for the interested reader, we also provide a larger batch of random samples in folder \emph{C.2.$\slash$all},
with the following levels of truncation: extreme truncation, where we sample all videos with the fixed zero latent representation; moderate truncation with $0.8$ threshold; and no truncation at all.

\subsection{Analysis of success and failure cases}
\label{ap-sec:success-failure}
 
As a result of the findings described in Section~\ref{ap-sec:latent-repr}, we find that our large-scale model also benefits from the truncation trick, in terms of generation quality. Focusing on the constant zero latent representation samples, overall, we find that our final model TRiVD-GAN-FP generates highly plausible motion patterns, including in highly complex cases such as intricate hand motions or water motion patterns. The model also makes plausible predictions of complex motion in a number of cases, where points move with a non-uniform acceleration across frames,
suggesting that it has learned a good representation for physical phenomena such as gravity (eg. for people jumping or juggling) and for some forms of object interactions (eg. for demonstrations of knitting). We highlight some of these in the accompanying video \emph{C.3.$\slash$annotated$\_$success$\_$constantz}.

We also highlight some failure cases in \emph{C.3.$\slash$annotated$\_$failure$\_$constantz$\_$vs$\_$notruncation}, where we provide constant zero latent representation samples on the left, and non-truncated samples on the right.
We find that when the truncation trick is not used, the model sometimes predicts very large motion, that leads to important object deformations. 
We also sometimes observe large artifacts across the entire scene.
Occasionally, the model makes predictions of obviously implausible trajectories.
In contrast, when using truncation, we see that these failures cases largely disappear. 

Across both levels of truncation however, 
we find that the model struggles when globally coherent motion must be predicted for objects with fine spatial structure, eg. in the case of hoola loops, or horse legs. 
We also find that in certain cases, motion seems to simply stop, presumably since this is an easy way for the model to yield plausible predictions in an overly conservative manner.
\amended{We note however that a trivial \emph{copy} baseline, which outputs a copy of the last input frame for each time step, performs very poorly at 330 on the Kinetics validation set - showing that consistently predicting trivial motion patterns is heavily penalized by the FVD.}
On rare occasions, we also observe objects disappearing into the void.
A final failure case comes up when the input sequence is blurry, which leads to poor predictions.

\section{Experimental Details}

\amended{

\subsection{Loss}
\label{ap-sec:loss}

The discriminator and generator are jointly trained with a geometric hinge loss~\citep{lim2017geometric}, updating the discriminators two times for each generator update. Specifically, the discriminator optimizes the following empirical risk:

\begin{equation}
    \frac{1}{D} \sum_{d=1}^D \rho(1 + \mathcal{D}(\mathcal{G}(z))) + \rho(1 - \mathcal{D}(x)),
\end{equation}

where $D$ is the discriminator output dimension, $\rho$ is the ReLU function, and $z$ and $x$ represent respectively a latent sample and a video. The generator optimizes the following empirical risk:

\begin{equation}
    - \frac{1}{D} \sum_{d=1}^D D(G(z)).
\end{equation}

Following DVD-GAN, we employ a variant of the projection conditioning of the discriminator \footnote{Communicated to us privately by the authors to help reproduce their results.},
which we call \emph{mixed projection discriminator}. 

We first recall the formulation of the original projection discriminator \cite{miyato2018cgans}.
Calling $h$ a batch of spatio-temporal features extracted from the videos fed to the discriminator, the projection discriminator maps these, on the one hand, to per-sample $r_x^b$ scalars only dependent on the input videos. On the other, a dot product is taken between $h$ and a learned vector, corresponding to a dummy zero class $y$ (or to the conditioning class if appropriate), to obtain a per-sample $r_{x,y}^{b}$ scalar dependent on both the videos and the class.
Next, these scalars are added together to form the discriminator output $o_b$, consisting of a single scalar per sample $b$.

Instead, the mixed projection discriminator extracts scalars $r_x^{b,t}$ and $r_{x,y}^{b,t}$ for each batch sample $b$ and each time step $t$ and computes $D=B \times B \times T$ outputs, where $B$ represents the batch size and $T$ the time dimension of $h$, obtained as follows:

\begin{equation}
    o_{b, b', t} = r_x^{b,t} + \sum_{t'} r_{x,y}^{b', t'}.
\end{equation}

We find that this variant improves over the original projection discriminator from 64.2 to 55.0 on the Kinetics validation set, for medium-sized models using a channel multiplier of 48 trained for 700k steps. As a result we employ it to train all our models.

\subsection{Implementation Details}
\label{ap:train-app}

\amended{
We preprocess the data as follows. Real video clips are first resized so that their smallest dimension is $64$, then randomly cropped and finally scaled to the range [-1, 1]. For evaluation, real and generated videos are then resized to resolution $224 \times 224$ using bilinear interpolation, for input to the I3D network.
}

Spectral Normalization~\citep{zhang2018self} is applied to all weights \amended{in the generator,} approximated by only the first singular value. 
All weights are initialized orthogonally~\citep{saxe2013exact}, and during training we maintain an auxiliary loss on the generator which penalizes weights diverging from orthogonality.
This loss is added to the GAN loss before optimization with weight $0.0001$.
The model is optimized using Adam~\citep{kingma2014adam} with batch size $512$, $\beta_1$ set to $0$, $\beta_2$ set to $0.999$ and a learning rate of $1\cdot 10^{-4}$ and $5\cdot 10^{-4}$ for $\gen$ and $\disc$ respectively.

All samples for evaluation (but not for training) use an exponential moving average of $\gen$'s weights, which is accumulated with decay $\gamma = 0.9999$ and is initialized after 20,000 training steps with the value of weights at that point. In order to disambiguate the effect of the weights' moving averages from those present in the Batch Normalization layers, we always evaluate using "standing statistics", where we run a large number of forward passes of the model (100 batches of 256 samples), and record the average means and variances for all Batch Normalization layers. These averages are used as the "batch statistics" for all subsequent evaluations. This process is repeated for each evaluation.

Conditioning of the Batch Normalization layers~\citep{ioffe2015batch, de2017modulating, dumoulin2017learned} is done by having the scale and offset parameters be an affine transformation of the conditioning vector. 
We note that the final batch normalization layer in $\gen$ is not conditional, and uses a regular learned scale and offset.
}

\amended{
\subsection{Estimating average step duration and memory consumption}
\label{ap:train}

We note that for calculating step time and maximum memory usage in Table 1 of the main paper, each training step corresponds to two $\disc$ updates and a single $\gen$ update, involving three passes of $\gen$ with batch size 8 and two passes of $\disc$ with batch size 16 alongside one with batch size 8. Furthermore, we emphasize that TPUs rely on the XLA intermediate compilation language, which automatically performs some re-materialization and optimization. As such, these numbers might change for other implementations.
}

\section{Further details on the recurrent units}

\subsection{Implementation details}

For the ConvGRU modules in DVD-GAN-FP, \citet{clark2019dvdgan} use ReLU activation functions, rather than the hyperbolic tangent (tanh).
For the ConvLSTM module, we experiment with both tanh, which is traditionally used, and the ReLU activation.
We find that ReLU significantly outperforms tanh early on, reaching 51.8 train FVD after 180K steps, against 62.7 for tanh. As a result, for our comparison of recurrent units, we report results using the ReLU activation function.

The hyper-network $f$ used to predict the parameters of our transformations differs depending on the output dimension.
In the case of PTSRU, we use a resolution-preserving, 2-hidden layer CNN.
For both TSRU and K-TrajGRU, a first subnetwork consists of a single convolutional layer, followed by adaptive max-pooling, of output resolution $4\times4$, then by a single hidden-layer MLP, to predict the set of kernels $\mathbf{w}$ used across the input.
The pseudo-segmentation map $\mathbf{S}$ is obtained using a single convolutional layer applied to the input.
For all three architectures, ReLU activations are interleaved between the linear layers, and we use a softmax activation on the output.
Each architecture has zero or two hidden layers, and we set the number of output hidden channels (or units in case of linear layers) to half the number of input channels.

\subsection{K-TrajGRU}

Our proposed K-TrajGRU is a kernel-based extension of TrajGRU \citep{shi2017neurips}, which we further detail here.
Formally, it is described by the following equations:

\begin{align}
    \centering
    &\theta_{h, x} = f(h_{t-1}, x_t), \label{eq:trajgru:th} \\
    &\tilde{h}_{t-1} = w(h_{t-1} ; \theta_{h, x}),  \label{eq:trajgru:wh} \\
    &r = \sigma(W_{r} \star_k [\tilde{h}_{t-1}; x_t] + b_r), \label{eq:trajgru-r} \\
    &h^{\prime}_t = r \odot \tilde{h}_{t-1}, \label{eq:trajgru-hp} \\
    &c = \rho(W_{c} \star_k [\tilde{h}^{\prime}_t ; x_t] + b_c), \label{eq:trajgru-c} \\
    &u = \sigma(W_{u} \star_k [\tilde{h}_{t-1}; x_t] + b_u), \label{eq:trajgru-u} \\
    &h_t = u \odot h_{t-1} + (1 - u) \odot c, \label{eq:trajgru-h}
\end{align}

where 
$h_{t-1}$ and $x_t$ denote respectively the past hidden features and the input features;
$f$ and $w$ are respectively the shallow convolutional neural network and the factorized warping function presented in the main paper and used by our TSRU modules;
$\star_k$ denotes a convolution operation with kernel size $k$,
$\rho$ is the activation function used (in our case, tanh),
$\sigma$ denotes the sigmoid function.
$\odot$ denotes elementwise multiplication,
and $W_{o}$ and $b_{o}$ are kernel and bias parameters for the convolutions used to produce each intermediate output $o$.

Conceptually, we see that K-TrajGRU and TSRU are related, in that they both warp the past hidden features to account for motion.
TSRU however has a simpler design, which can be intuitively interpreted: it fuses the warped features $\tilde{h}_{t-1}$ and novelly generated content $c$, using a predicted gate $u$.
K-TrajGRU, instead, provides the warped features to equations (\ref{eq:trajgru-r}, \ref{eq:trajgru-c}, \ref{eq:trajgru-u}), but still combines the input hidden features $h_{t-1}$ unchanged, together with $c$.

We note that \citet{shi2017neurips} propose to predict $L$ warped versions of the hidden state, and that our proposed recurrent unit specifically extends the case where $L=1$.
Increasing $L$, as well as the addition of a read gate to the TSRU module, are potential ideas for further improving our transformation-based recurrent unit's expressivity, which we leave to be explored in future work.

\subsection{ConvLSTM}

For comparison with other recurrent units, the hidden state and memory cell of our ConvLSTM baseline each have half as many channels as the hidden state used in the other designs, that do not maintain a memory cell (ConvGRU, K-TrajGRU and TSRU variants).
The hidden state and the memory cell are initialized with respectively the first and the second half of the conditioning sequence's encoding.
Like for the other recurrent units, this encoding is obtained by passing the concatenated image-level features extracted by the encoder for each frame, through a 3x3 convolution layer, followed by a ReLU activation, to compress the representation to the right dimension.
Then, the output hidden state and memory cell for each time step are concatenated and passed on to the subsequent residual blocks of the generator.

\subsection{Dynamic ConvLSTM}
\label{ap:dynamic-convlstm}

\amended{
The dynamic ConvLSTM proposed by \citet{xu2018cvpr} is conceptually simple: for each sample, a set of kernels are predicted and used in the convolutional layers applied either on the recurrent unit's input features $x$ or on its hidden features $h$, in the input, forget, output gate equations, as well as for the intermediate state's equation - totaling 8 predicted kernels.}
The network predicting these kernels, denoted by ``Kernel CNN'', employs a channel-wise softmax operation along the input channel to increase sparsity of the predicted kernels.
We refer the reader to the Section 3.3 of the original paper for a more extensive presentation.

It is important to note that such dynamic convolutions require a different $2D$ kernel to be predicted for each input and output channel.
To keep to a small number of parameters of the Kernel CNN,
\citet{xu2018cvpr} propose to share weights across the input and output channel dimensions, by mapping the difference between image features extracted from the two previous frames 
$F_t[i] - F_{t-1}[j]$ to the corresponding 2D kernel $W(i, j)$, where $F[i]$ denotes the $i-th$ channel of features $F$, 
and where $W(i, j)$ corresponds to the filter used for input channel $i$ and output channel $j$.
This relies on the fact that the generator is autoregressive, and encodes each prediction before making the next one.
We adapt this to our architecture, as described later in this section for the sake of completeness.

Before going into more implementation details however, we draw the attention of the reader to the following important analysis.
In contrast with dynamic convolutions, our proposed transformation-based recurrent units all essentially rely on dynamic, depth-wise, locally connected layers.
Hence, while these recurrent units predict transformation parameters whose dimension scales with the spatial dimensions of the hidden features, 
the dimension of the Dynamic ConvLSTM transformation's parameters scales with the number of input and output channels.
We find that this leads to a largely increased computational cost, even for medium sized models.
Specifically, with extensive use of parallelisation across devices, when using Dynamic ConvLSTMs, 
our medium-size models, that use a channel multiplier of 48,
run at an average step duration of 6.385s for a batch size of 512;  while in the same set up, models using ConvGRU (resp. TSRU) run at 284ms (resp. 354ms) per step.
In this context, the Dynamic ConvLSTM module hence only allows a very limited width of architectures.
In this sense, we argue that it is not scalable, and therefore we do not consider it for comparison with other recurrent units for the purposes of our large-scale setting.

\paragraph{Implementation details} 
To allow the use of their proposed recurrent unit in our architecture, 
at time step $t$, we treat the first $c_h$ channels of the previous and current features $(x_{t-1}, x_{t})$ as the input features to the Kernel CNN,
where $c_h$ denotes the number of channels of the hidden representation of the recurrent unit.
We note that this requires the number of channels $c_x$ of the inputs features $x$ to be a multiple of $c_h$, so that each difference of features can be mapped to $q+1$ kernels, where $q$ is such that $c_x = q \times c_h$.
For the first time step, we use the first $c_h$ channels of the sequence encoding to play the role of the previous features $x_{t-1}$.

A second difference is that their proposed unit additionally fuses the past hidden features, and past cell states to produce the hidden and cell states that are input to the module. 
This operation is denoted by $Fus-4$ in the paper.
This might be redundant with the intended function of the hidden state and memory cell of a ConvLSTM. In practice, it is observed by the authors to bring only a marginal gain in general.
Additionally, it is an orthogonal contribution, which could be implemented for any of the other recurrent units we have considered.
As a result, we do not implement this aspect of their recurrent unit.

\amended{
\subsection{TSRU variants}

We provide in Figure~\ref{fig:more-tsru-variants} an illustration for all proposed TSRU variants described in Section 3.2.

\begin{figure*}[t]
\centering
\includegraphics[width=1.0\textwidth]{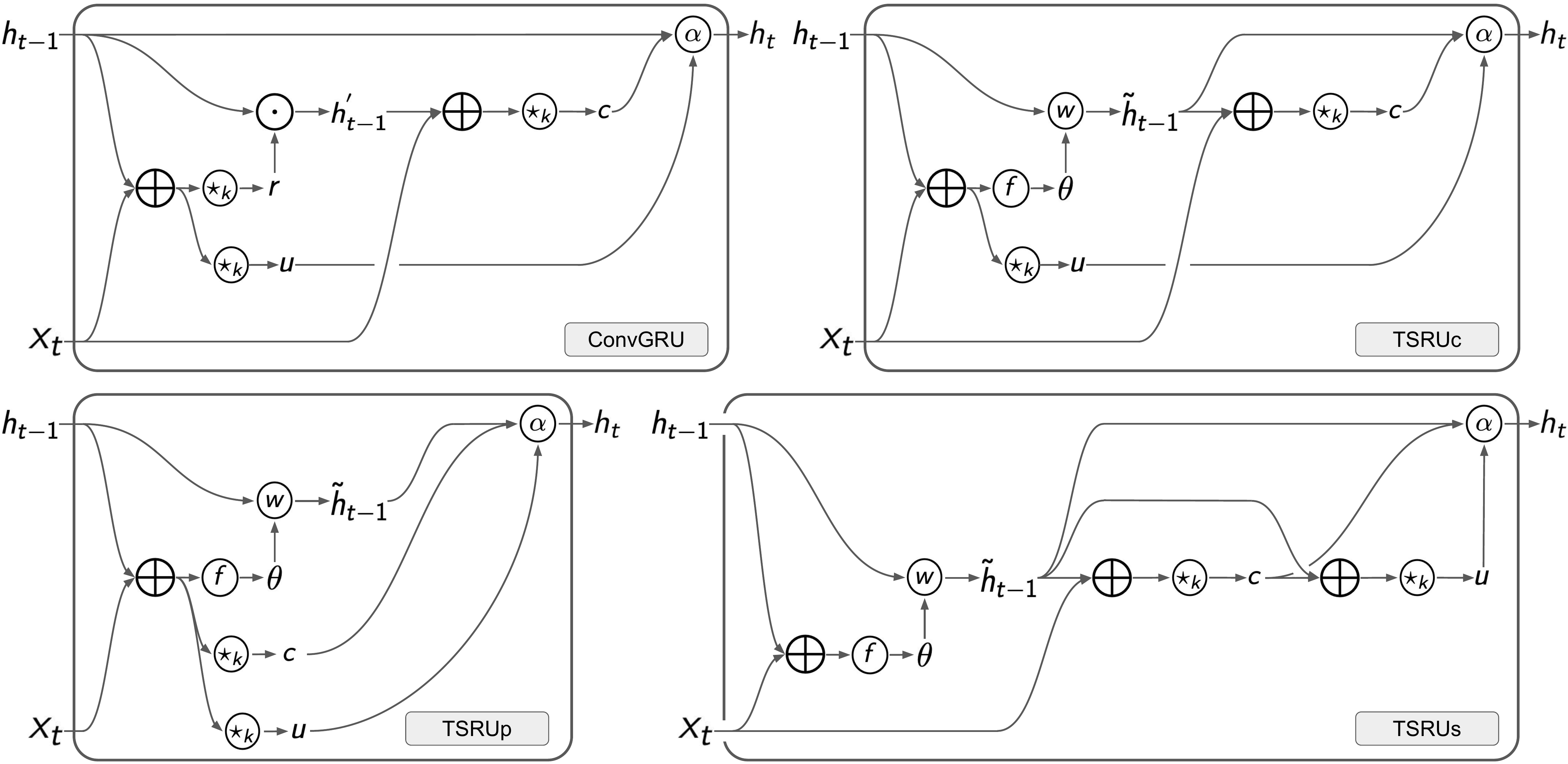}
\vspace{-1em}
\caption{\amended{Information flow for ConvGRU (top-left) and our proposed \tsru{}$_c$ (top-right), \tsru{}$_p$ (bottom-left) and \tsru{}$_s$ variants (bottom-right); where $\alpha$ represents elementwise convex combination of $h_{t-1}$ and $c$ with coefficient provided by $u$; $\star_k$, convolution with kernel size $k$; $\bigoplus$, concatenation and $\bigodot$, elementwise multiplication.
}}
\label{fig:more-tsru-variants}
\end{figure*}

}

\end{document}